\documentclass{article}
\usepackage{ijcai24}

\usepackage{times}
\usepackage{soul}
\usepackage{url}
\usepackage{multirow}
\usepackage[hidelinks]{hyperref}
\usepackage[utf8]{inputenc}
\usepackage[small]{caption}
\usepackage{graphicx}
\usepackage{amsmath}
\usepackage{amsthm}
\usepackage{amsfonts}
\usepackage{booktabs}
\usepackage{algorithm}
\usepackage{algorithmic}
\usepackage[switch]{lineno}

\title{Text-to-Image GAN with Pretrained Representations}

\author{
Xiaozhou You$^1$
\and
Jian Zhang$^1$\\
\affiliations
$^1$School of Electronic and Computer Engineering, Peking University
\\
\emails
youxz@stu.pku.edu.cn,
zhangjian.sz@pku.edu.cn
}

\begin{document}

\maketitle

\begin{abstract}
    Generating desired images conditioned on given text descriptions has received lots of attention. Recently, diffusion models and autoregressive models have demonstrated their outstanding expressivity and gradually replaced GAN as the favored architectures for text-to-image synthesis. However, they still face some obstacles: slow inference speed and expensive training costs. To achieve more powerful and faster text-to-image synthesis under complex scenes, we propose TIGER, a text-to-image GAN with pretrained representations. To be specific, we propose a vision-empowered discriminator and a high-capacity generator. (i) The vision-empowered discriminator absorbs the complex scene understanding ability and the domain generalization ability from pretrained vision models to enhance model performance. Unlike previous works, we explore stacking multiple pretrained models in our discriminator to collect multiple different representations. (ii) The high-capacity generator aims to achieve effective text-image fusion while increasing the model capacity. The high-capacity generator consists of multiple novel high-capacity fusion blocks (HFBlock). And the HFBlock contains several deep fusion modules and a global fusion module, which play different roles to benefit our model. Extensive experiments demonstrate the outstanding performance of our proposed TIGER both on standard and zero-shot text-to-image synthesis tasks. On the standard text-to-image synthesis task, TIGER achieves state-of-the-art performance on two challenging datasets, which obtain a new FID 5.48 (COCO) and 9.38 (CUB). On the zero-shot text-to-image synthesis task, we achieve comparable performance with fewer model parameters, smaller training data size and faster inference speed. Additionally, more experiments and analyses are conducted in the Supplementary Material.
\end{abstract}

\begin{figure}[!t]
	\setlength{\abovecaptionskip}{0.1cm}
	\centerline{\includegraphics[width=1\linewidth]{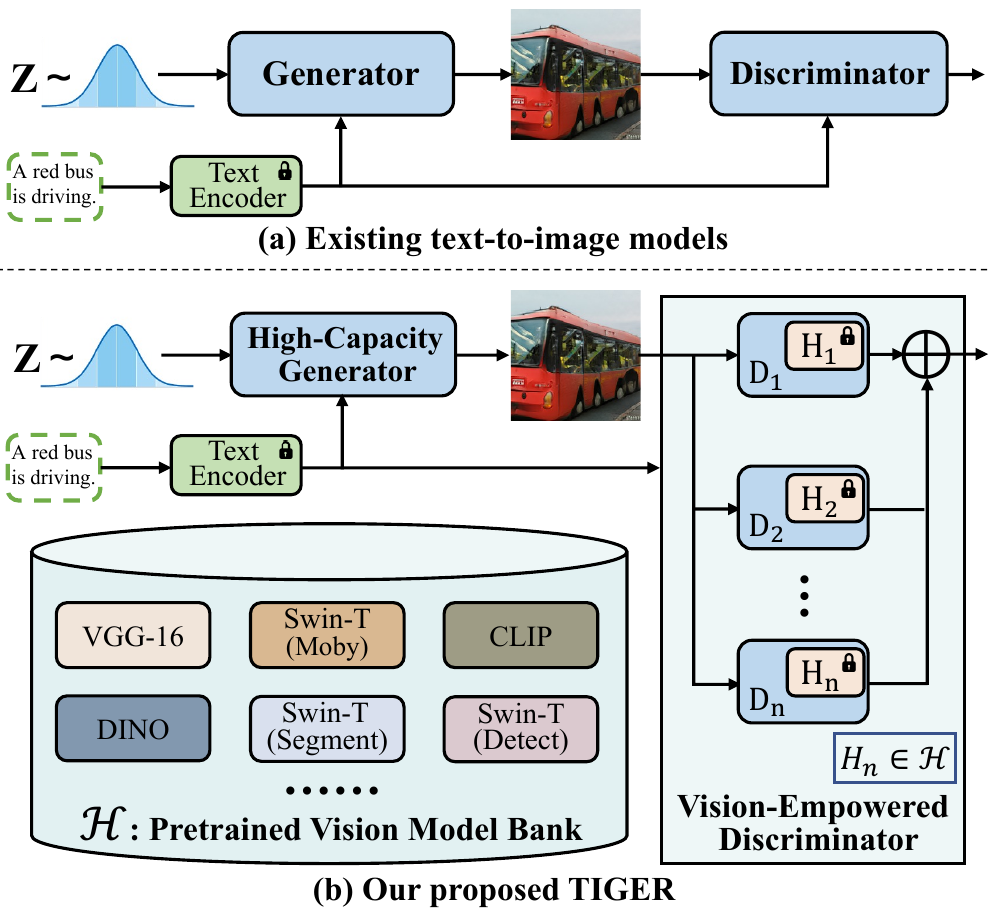}}
	\caption{(a): Existing text-to-image GANs were trained from scratch. (b): To build a more powerful and faster GAN, our proposed TIGER consists of a vision-empowered discriminator and a high-capacity generator. The vision-empowered discriminator consists of several sub discriminator $D_{i}$, which contains the model $H_{i}$ selected from the pretrained vision model bank $\mathcal{H}$ to enhance the complex scene understanding ability and domain generalization ability. And our high-capacity generator can achieve effective cross-modal text-image fusion while increasing model capacity.}
	\label{nn1}
\end{figure}
\begin{figure*}[!t]
	\setlength{\abovecaptionskip}{0.1cm}
	\centerline{\includegraphics[width=0.82\linewidth]{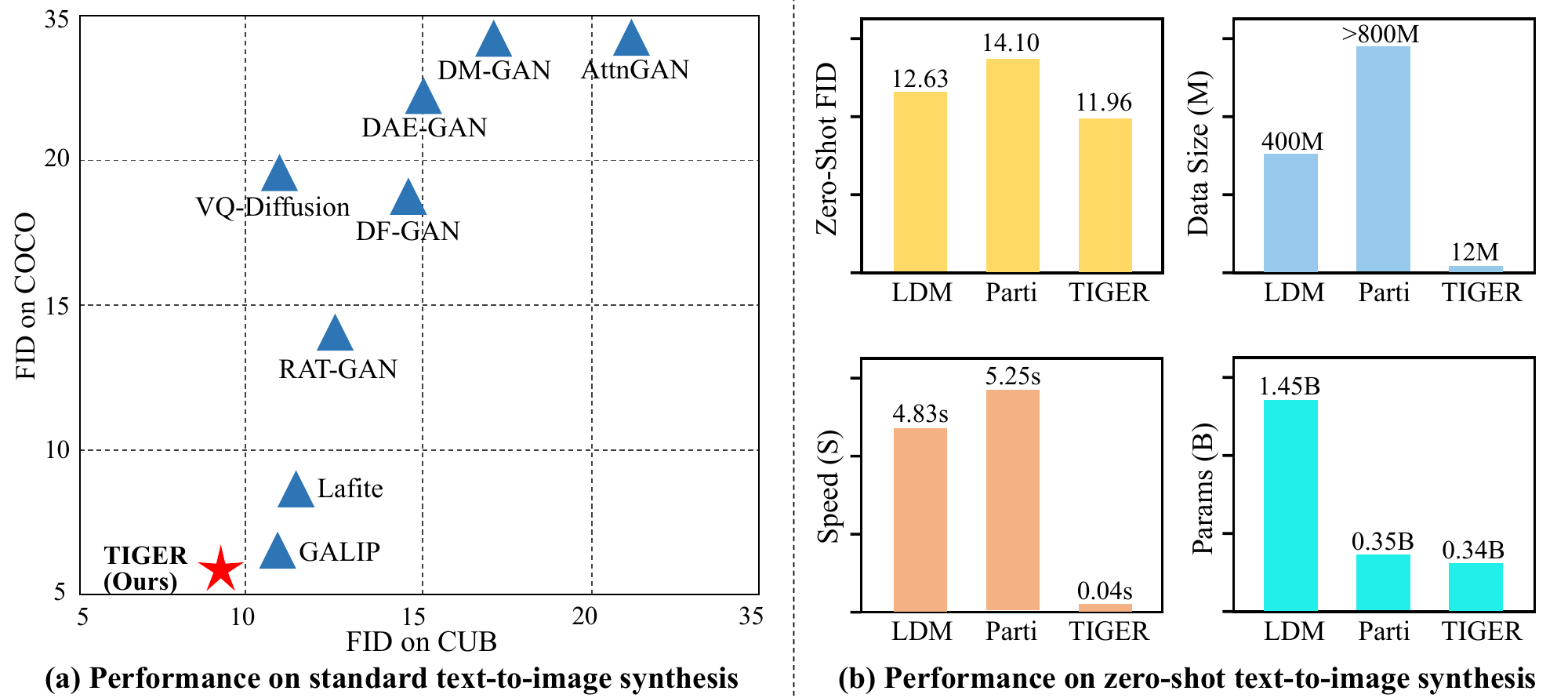}}
	\caption{The outstanding performance of our proposed TIGER. (a): On the standard text-to-image synthesis task, our TIGER achieves a new state-of-the-art FID 5.48 (COCO) and 9.38 (CUB). Lower FID means better. (b): On the zero-shot text-to-image synthesis task, our TIGER achieves comparable performance (zero-shot FID) with fewer model parameters, smaller training data size and 120$\times$ faster inference speed than LDM (DF) and Parti-350M (AR).}
	\label{nn2}
\end{figure*}
\begin{figure*}[t]
	\setlength{\abovecaptionskip}{0.1cm}
	\centerline{\includegraphics[width=0.9\linewidth]{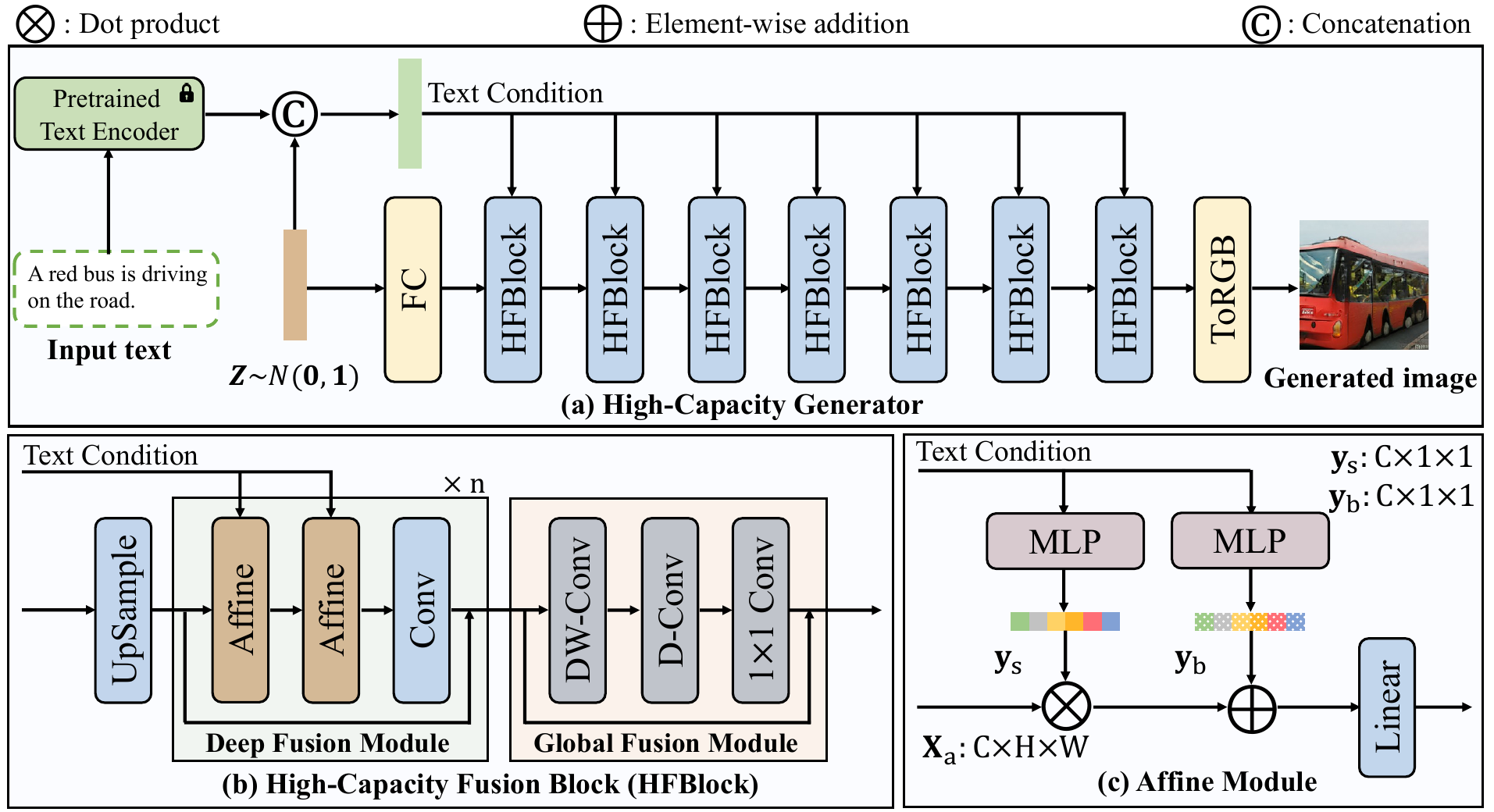}}
	\caption{The architectures we investigate. (a): The high-capacity generator consists of several high-capacity fusion blocks to generate desired images under complex scenes. (b): The high-capacity fusion block includes several deep fusion modules and a global fusion module to further improve model capacity, in which ``D-Conv'' stands for dilated convolutional network. (c): The affine module can achieve effective cross-modal text-image fusion.}
	\label{nn3}
\end{figure*}

\section{Introduction}

Recently, Artificial Intelligence Generated Content (AIGC) has become a hot topic in the academic community. Among them, text-to-image synthesis is one of the most attractive areas. Among them, text-to-image synthesis is one of the most attractive areas due to its potential value in many fields, such as computer-aided design, virtual scene generation, and photo editing. It aims to generate desired images conditioned on given text descriptions. To reach this, many methods have been proposed, including generative adversarial networks \cite{xu_attngan_2018}, diffusion models \cite{gu2022vector}, variational auto-encoders \cite{kingma_auto-encoding_2014}, autoregressive models \cite{ramesh2021zero}, \textit{etc.} In the early years, GAN was the main tool to realize text-to-image synthesis.

Due to GAN's unstable training and weak performance under complex scenes, some researchers try to adopt diffusion models (DF) and autoregressive models (AR) to accomplish the task. Based on large-scale datasets and massive parameters, they have demonstrated outstanding expressivity and gradually replaced GAN as the favored architecture for text-to-image synthesis. However, they still suffer from two limitations. First, large-scale models have slow inference speed. The progressive denoising of diffusion models and the token-by-token generation of autoregressive models make their speed far behind GAN \cite{tao2023galip}. For example, to generate an image, GLIDE (DM) costs 15s, Parti (AR) costs 5s, and Lafite (GAN) only costs 0.02s. Second, the training of large-scale models requires massive training data and parameters, which brings huge training burden and computing requirements. For example, DALL-E \cite{ramesh2021zero} has 12B parameters and is trained on 250M image-text pairs. These characteristics seriously hinder their further application.

To address them, we revisit GAN. Compared with diffusion model and autoregressive model, GAN has faster inference speed but weaker performance under complex scenes. Pointed out by some works \cite{kumari2022ensembling,sauer2021projected,tao2023galip}, pretrained vision models have better complex scene understanding and domain generalization capabilities, such as DINO \cite{caron2021emerging}, CLIP \cite{radford2021learning}, Swin \cite{liu2021swin}. Motivated by this, we decided to introduce representations from pretrained vision models into GANs.


In this work, we propose TIGER, a \textbf{t}ext-to-\textbf{i}mage \textbf{G}AN with pr\textbf{e}trained \textbf{r}epresentations, which aimed to build a more powerful and faster text-to-image model. To be specific, we propose a vision-empowered discriminator and a high-capacity generator. The vision-empowered discriminator absorbs the complex scene understanding ability and the domain generalization ability from pretrained vision model to enhance the performance under complex scenes. Unlike previous works \cite{sauer2021projected,tao2023galip}, we explore to collect multiple representations from different models, which means we stack multiple pretrained models in our discriminator. And we freeze the parameters of pretrained vision models in order to maintain their original characteristics. Besides, we propose a high-capacity generator, aiming to achieve effective text-image fusion while increasing the model capacity. The high-capacity generator consists of multiple novel high-capacity fusion blocks. And the high-capacity fusion block contains several deep fusion modules and a global fusion module. The goal of the deep fusion module is to achieve efficient text-image fusion, and the goal of the global fusion module is to enhance model expressivity both on spatial and channel dimensions. 

As shown in Fig. \ref{nn2}, extensive experiments demonstrate the outstanding performance of TIGER both on standard and zero-shot text-to-image synthesis tasks. On the standard text-to-image synthesis task, TIGER achieves state-of-the-art performance on two challenging datasets. On the zero-shot text-to-image synthesis task, we achieve a comparable performance with fewer model parameters, smaller training data and 120$\times$ faster inference speed than LDM \cite{rombach2022high} and Parti-350M \cite{yu2022scaling}.


\section{Related Works}
\subsubsection{Text-to-Image Synthesis.}
Reed \textit{et al.} first proposed employing conditional generative adversarial networks to generate images under text conditions in 2016 \cite{reed_learning_2016}, which opened the door to text-to-image synthesis. To further improve the quality of generated images, Zhang \textit{et al.} proposed stacking multiple generator-discriminator pairs to gradually generate high-quality images from coarse to fine under text conditions \cite{zhang_stackgan_2017}. During training, multiple generator-discriminator pairs are required to coordinate to generate high-quality images. After that, Xu \textit{et al.} proposed AttnGAN \cite{xu_attngan_2018} to achieve word-level fine-grained generation by introducing a word-level attention mechanism. To overcome the limitations of stacked architecture, Ming \textit{et al.} proposed DF-GAN \cite{tao2020df}, which aims to employ single generator to complete text-to-image synthesis. In addition, he also proposed utilizing MA-GP to generate text-matching images. The following SSA-GAN \cite{liao2022text} and RAT-GAN \cite{ye2022recurrent} also adopted single-stage architecture. SSA-GAN proposes semantic-spatial condition batch normalization, which employs mask map to overcome the problem of insufficient spatial fusion in DF-GAN. RAT-GAN proposes Recurrent Affine Transformation to model long-range dependencies between fusion blocks. 


\subsubsection{Large-Scale Text-to-Image Models.}
Recently, based on large-scale pre-training, diffusion models and autoregressive models have started to show their influence on zero-shot text-to-image synthesis. DALL-E \cite{ramesh2021zero} is the first large-scale zero-shot text-to-image model, which trains a 12B Transformer based on autoregressive model. Similar to DALL-E, CogView \cite{ding2021cogview} trains a 4B parameter autoregressive model and finetunes it for various downstream tasks. Then, more large-scale text-to-image generative models are proposed, further demonstrating their expressiveness, such as CogView2 \cite{ding2022cogview2}, Muse \cite{chang2023muse}, GLIDE \cite{nichol2021glide}, and eDiff \cite{balaji2022ediffi}. The Parti \cite{yu2022scaling} attempts to treat text-to-image synthesis as a translation task, which proposes a sequence-to-sequence autoregressive model. Latent Diffusion Models (LDM) \cite{rombach2022high} introduce the latent space to diffusion models to enable the training on limited computational resources.

\begin{figure*}[t]
	\setlength{\abovecaptionskip}{0.1cm}
	\centerline{\includegraphics[width=0.88\linewidth]{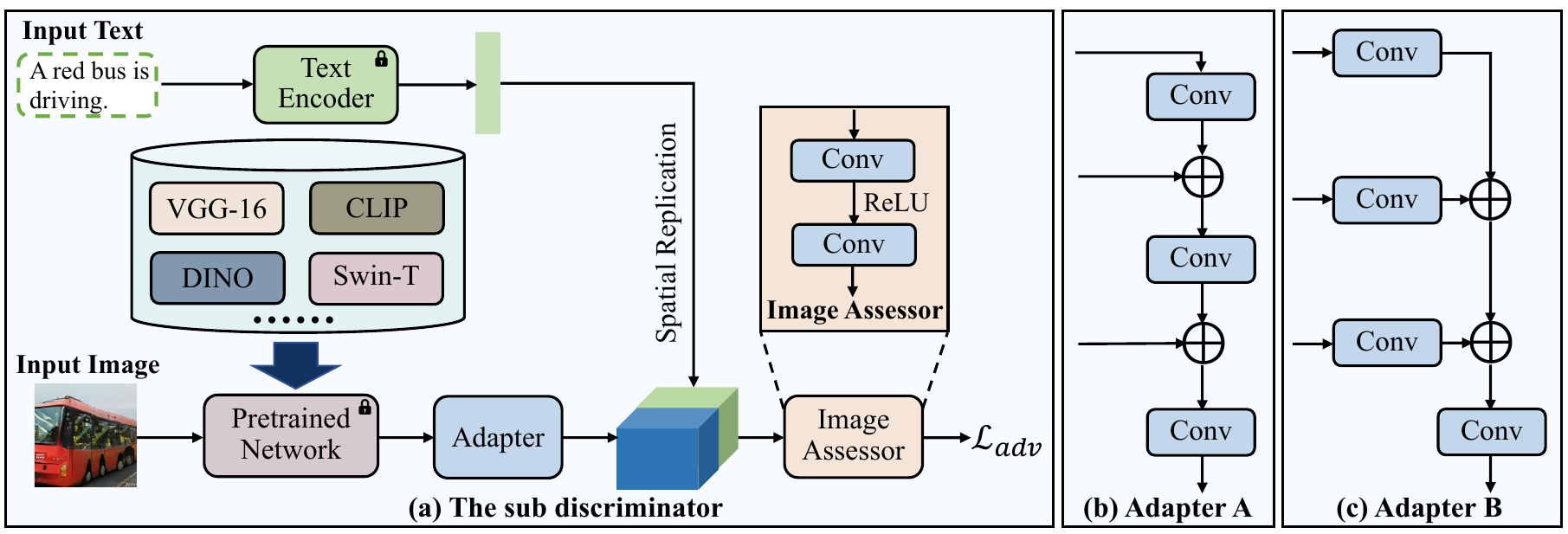}}
	\caption{(a): The architecture of sub discriminator in our vision-empowered discriminator. In our vision-empowered discriminator, each sub discriminator processes the representation from different pretrained vision networks. (b) and (c):  The different architectures of adapter. For multi-level features, we try two different adapters to enhance model performance.}
	\label{nn45}
\end{figure*}

\section{Methods}
In this paper, we propose TIGER, a text-to-image GAN with pretrained representations, which aimed to build a more powerful and faster text-to-image model. We will introduce the novel components of TIGER in this section, including: high-capacity fusion block and vision-empowered discriminator.

\subsection{Model Overview}
The overall architecture of TIGER is shown in Fig. \ref{nn1}. Unlike previous works \cite{sauer2021projected,tao2020df}, we explore introducing multiple pretrained vision models in our network, and the goal is to utilize the representations from pretrained models to further improve model performance. Specifically, we propose an innovative vision-empowered discriminator and an innovative high-capacity generator. The vision-empowered discriminator consists of several sub discriminators, each of which processes the extracted representation from different pretrained vision model. The overall adversarial loss of the vision-empowered discriminator is the sum of sub discriminators' loss. The architecture of the sub discriminator is shown in Fig. \ref{nn45} (a).

The architecture of the high-capacity generator is shown in Fig. \ref{nn3} (a). In order to increase the model capacity and achieve effective cross-mode text-image fusion, we propose a novel high-capacity fusion block (HFBlock) in our proposed high-capacity generator. First, we generate a 100-dimensional random noise sampled from the Gaussian distribution. And the random noise is used as the input of first layer. Then, the text descriptions are encoded to get sentence vector by a pretrained text encoder. We concatenate random noise with sentence vectors as text condition, which is the input to the high-capacity fusion block.

\subsection{High-Capacity Fusion Block}
As shown in Fig. \ref{nn3} (b), the High-Capacity Fusion Block consists of several deep fusion modules and a global fusion module. The goal of the deep fusion module is to achieve efficient cross-modal text-image fusion, and the goal of the global fusion module is to enhance the model expressivity both on spatial and channel dimensions. We describe them in detail next.

\subsubsection{Deep fusion Module.} To achieve effective cross-modal text-image fusion, we introduce several deep fusion modules into our network \cite{tao2020df}. Compared with the previous attention-based method \cite{xu_attngan_2018}, the deep fusion module has a lower computational cost, which can benefit the training of model. Compared with the previous concatenation method \cite{reed_generative_2016}, the deep fusion module can more effectively complete the text-image fusion \cite{liao2022text}. Specifically, the deep fusion module consists of two affine modules and a $3\times3$ convolutional neural network. The architecture of affine module is shown in Fig. \ref{nn3} (c). After the affine module, we employ a ReLU function to add nonlinearity to fusion process. And we stack two deep fusion modules in our HFBlock ($n = 2$ by default).

\subsubsection{ Global fusion Module.} Since the afﬁne module only fuses the text condition separately for each channel, it lacks information fusion across channel dimensions. To enhance the model capacity both on spatial and channel dimensions, we propose the global fusion module \cite{guo2022visual}. As shown in Fig. \ref{nn3} (b), the global fusion module consists of a $3\times3$ depth-wise convolutional network, a $5\times5$ dilated depth-wise convolutional network with dilation being 3, and a $1\times1$  convolutional network. The $3\times3$ depth-wise convolutional network focuses on modeling and enhancing information fusion on the spatial dimensions. The $5\times5$ dilated depth-wise convolutional network is aimed to extract long-range dependencies. And the $1\times1$  convolutional network is employed to facilitate cross-modal fusion and explicitly model the inter-dependencies between channels.

\subsection{Vision-Empowered  Discriminator}

As pointed out by previous works \cite{sauer2021projected,kumari2022ensembling}, GAN has weak performance under complex scenes and faces some obstacles, such as the lack of diversity and unstable training. To build a powerful GAN, we propose a vision-empowered discriminator, in which we explore introducing multiple representations from different pretrained vision models. These models are pretrained on huge and complex datasets, which include many different scenarios \cite{radford2021learning,caron2021emerging}. We argue that these pretrained model has outstanding complex scene understanding ability and domain generalization ability \cite{tao2023galip}, which helps to enhance the performance of GAN under some complex datasets.

Unlike previous works, we integrate multiple pretrained representations in the discriminator instead of one. We believe that integrating multiple pretrained representations from different vision networks in the discriminator can be beneficial to improve performance. Specifically, the vision-empowered discriminator consists of multiple sub discriminators, each of which mainly processed the representations from different pretrained vision models. The framework of the sub discriminator is shown in Fig. \ref{nn45} (a). For generated images or real images, we exploit the selected pretrained network to extract representation. Then, the extracted representation are processed by an adapter to obtain vision features. For single-level representation (such as VGG and Swin), we employ two sequential convolutional networks as adapter. For multi-level representation (such as CLIP and DINO), we try two different adapters to improve performance (Fig. \ref{nn45} (b) and (c)). The text description is encoded by a pretrained text encoder to obtain text features. Then, the replicated text features and vision features are concatenated together. Finally, the image assessor consisting of several convolutional networks evaluates the quality of the image. And its result is the adversarial loss of the sub discriminator. The total discriminator adversarial loss is the sum of all sub discriminator adversarial losses.

\subsection{Loss Function}
\subsubsection{Semantic Contrastive Loss.} To promote the semantic consistency between the generated image and text description, we introduce semantic contrastive loss in our model. Specifically, we compute the cosine similarity between text features and image features extracted by the CLIP model \cite{radford2015unsupervised}. The mathematical form is as follows:
\begin{equation}
	\begin{aligned}
		\mathcal{L}_{\text{CLIP}}=  \text{Cos}( \mathbf{I}, \mathbf{T} ),
	\end{aligned}
\end{equation}
where $\text{Cos}$ is the cosine function, $\mathbf{I}$ and  $\mathbf{T}$ are the encoded image features and text features extracted by the CLIP model, respectively.
\subsubsection{Overall Loss.}
For the i$^{th}$ sub discriminator in our vision-empowered discriminator, we use hinge loss with MA-GP \cite{tao2020df} as the sub discriminator loss $\mathcal{L}^{adv}_{D_{i}}$. The speciﬁc mathematical form is as follows:
\begin{equation}
	\begin{aligned}
		\mathcal{L}^{adv}_{D_{i}} &= \mathbb{E}_{\mathbf{x}\sim P_{data}}[\max(0,1-D_{i}(\mathbf{x},\mathbf{s}))]\\
		&\quad + \frac{1}{2}\mathbb{E}_{\mathbf{x}\sim P_{G}}[\max(0,1+D_{i}(\hat{\mathbf{x}},\mathbf{s}))]\\
		&\quad + \frac{1}{2}\mathbb{E}_{\mathbf{x}\sim P_{data}}[\max(0,1+D_{i}(\mathbf{x},\hat{\mathbf{s}}))],
	\end{aligned}
\end{equation}
where $\mathbf{x}$ is real image, $\hat{\mathbf{x}}$ is generated image, $\mathbf{s}$ is matched sentence, $\hat{\mathbf{s}}$ is unmatched sentence, $D_{i}$ is the i$^{th}$ sub discriminator, respectively. The overall adversarial loss $	\mathcal{L}_{D}$ of vision-empowered discriminator is the sum of the sub discriminator loss. The speciﬁc mathematical form is as follows:


\begin{equation}
	\begin{aligned}
		\mathcal{L}_{D} &= \sum_{i=1}^{M} \lambda_{i} \cdot \mathcal{L}^{adv}_{D_{i}},
	\end{aligned}
\end{equation}
where $ \lambda_{i} $ is is the corresponding hyperparameter, $M$ is the total number of sub discriminators, respectively. The training loss of generator $\mathcal{L}_{G}$ is as follows:
\begin{equation}
	\begin{aligned}
		\mathcal{L}_{G} &= \lambda_{\text{CLIP}} \cdot \mathcal{L}_{\text{CLIP}} + \mathcal{L}_{adv}^{G}, \\
		\mathcal{L}_{adv}^{G} &= -\sum_{i=1}^{M} \lambda_{i} \cdot \mathbb{E}_{\mathbf{x}\sim P_{data}}[D_{i}(\hat{\mathbf{x}},\mathbf{s})].
	\end{aligned}
\end{equation}
where $\lambda_{\text{CLIP}}$ is a hyper-parameter, respectively.

\begin{figure*}[t]
	\setlength{\abovecaptionskip}{0.15cm}
	
	\centerline{\includegraphics[width=\textwidth]{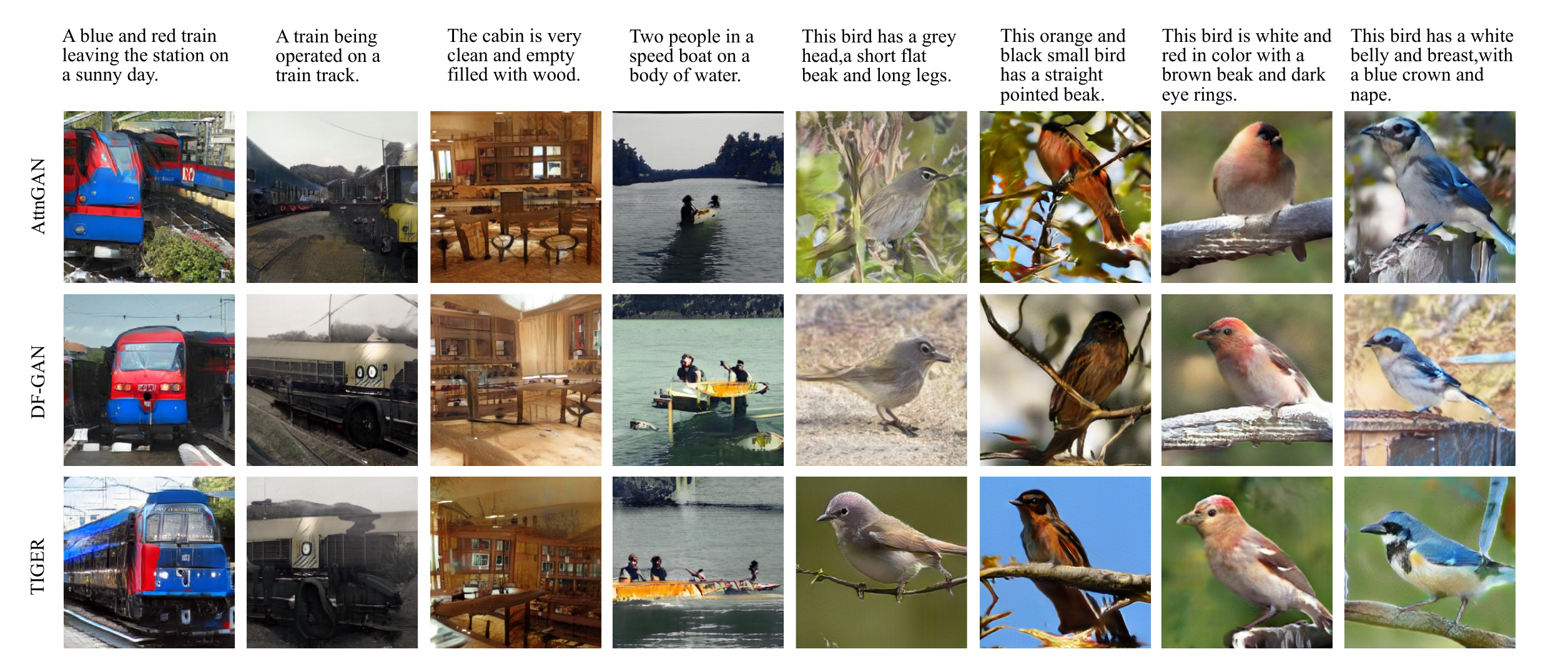}}
	\caption{Qualitative comparison between AttnGAN, DF-GAN, and our proposed TIGER conditioned on text descriptions from the test set of COCO datasets (1st - 4th columns) and CUB datasets (5th - 8th columns).}
	\label{nn5-1}
\end{figure*}
\section{Experiments}
In this section, we introduce the datasets, training details, and evaluation details. Then, we conduct our experiments to verify the effectiveness and superiority of our proposed TIGER.

\noindent{\bf Datasets.} We evaluate the proposed model on three challenging datasets, i.e., CUB bird \cite{wah2011caltech}, COCO \cite{lin2014microsoft} and CC12M \cite{changpinyo2021conceptual}. The COCO datasets contain 80k images for training and 40k images for testing. Each image has ﬁve language descriptions. Besides, the COCO dataset is always employed in recent works to evaluate the performance of complex image synthesis. The CUB bird datasets (200 categories) contain 8855 training images and 2933 testing images. Each image has 10 text descriptions. The CC12M datasets contain about 12 million text-image pairs,  and it's always adopted for pretraining and to evaluate the zero-shot performance of the text-to-image model.

\begin{table}[t] \small
	\centering
	\resizebox{1\linewidth}{!}{
		\begin{tabular}{l|c|c|c|c}
			\toprule
			\multirow{2}{*}{Model}                 & \multicolumn{2}{c|}{CUB}                            & \multicolumn{2}{c}{COCO}             \\ 
			\cline{2-5}   
			& FID $\downarrow$        & R-precision $\uparrow$             & FID $\downarrow$          & R-precision $\uparrow$   \\ \midrule
			AttnGAN &23.98 & 0.246 & 35.49 & 0.183\\
			DM-GAN              & 16.09                   & 0.287                        & 32.64                     & 0.236              \\
			XMC-GAN       & -                       & -                         & 9.30                      & -                  \\
			DAE-GAN        & 15.19                   & 0.321                         & 28.12                     & 0.257            \\
			RAT-GAN& 13.91 & 0.338 & 14.60& 0.298\\
			VQ-Diffusion     & 10.32                   & -                         & 13.86                     & -             \\
			GALIP                     & 10.08       & 0.325          & 5.85          & 0.342 \\
			LAFITE        & 14.58                   & 0.336                    & 8.21                      & 0.332         \\ 
			
			DF-GAN             & 14.81                   & 0.306                & 19.32                     & 0.278       \\ 
			\midrule
			\textbf{TIGER (ours)} & \textbf{9.38} & \textbf{0.341}&\textbf{5.48}& \textbf{0.348}\\
			\bottomrule
	\end{tabular}}
	\caption{The results of FID and R-precision compared with the state-of-the-art methods on the test set of CUB and COCO.}
	\label{table1}
	\vspace{-0.4cm}
\end{table}

\noindent{\bf Training Details.} Our model is implemented in PyTorch. The Adam optimizer \cite{kingma2014adam} with $\beta_{1}$ = 0.0 and $\beta_{2}$ = 0.9 is used in the training. The learning rate is set to $1\times10^{-4}$ for generator and $4\times10^{-4}$ for discriminator according to TTUR \cite{heusel2017gans}. The hyper-parameters $\lambda_{\text{CLIP}}$ is set to 4. And the proposed vision-empowered consists of 2 sub discriminators (CLIP, DINO). The corresponding hyper-parameters $\lambda_{1}$, $\lambda_{2}$ are set to 1 and 0.001, respectively. And we use the CLIP text encoder as our text encoder. 

\noindent{\bf Evaluation Details.} The Fr\'{e}chet Inception Distance (FID) \cite{heusel2017gans} and top-1 R-precision \cite{xu_attngan_2018} are used to evaluate the performance of our work. For FID, it computes the Fr\'{e}chet distance between the distribution of the generated images and real-world images in the feature space of a pre-trained Inception v3 network \cite{szegedy2016rethinking}. Lower FID means model achieves better performance. For R-precision, we use CLIP \cite{radford2021learning} to calculate the cosine similarity between original image and given description. Following previous works \cite{liao2022text,tao2022df}, we do not use IS \cite{salimans2016improved} on COCO and CUB datasets because it can't evaluate the image quality well. Moreover, we evaluate the number of parameters (Param), inference speed (Speed) and training data size (Data) to compare with current methods.

\subsection{Quantitative Evaluation}
As shown in Tab. \ref{table1}, we conduct the quantitative comparison between our proposed TIGER and state-of-the-art methods on standard text-to-image synthesis task, including: AttnGAN \cite{xu_attngan_2018}, DM-GAN \cite{zhu_dm-gan_2019}, XMC-GAN \cite{zhang_cross-modal_2021}, DAE-GAN \cite{ruan_dae-gan_202}, RAT-GAN \cite{ye2022recurrent}, DF-GAN \cite{tao2022df}, Lafite \cite{zhou_lafite_2021}, VQ-diffusion \cite{gu2022vector} and GALIP \cite{tao2023galip}. On COCO datasets, TIGER reaches a new state-of-the-art FID 5.48. Compared with the single-stage architecture baseline DF-GAN, TIGER decreases the FID metric from 19.32 to 5.48 and improves the R-precision metric from 0.278 to 0.348, which shows that our GALIP obtains a huge performance boost. And Compared with the diffusion model VQ-diffusion, TIGER decreases the FID metric from 13.86 to 5.48. Especially, compared with the recent Lafite, TIGER decreases the FID metric from 8.21 to 5.48 and improves the R-precision metric from 0.332 to 0.348, which demonstrates that integrating pretrained representations is effective. On CUB datasets, TIGER also achieves leading results. Compared with the recent GALIP, TIGER decreases the FID metric from 10.08 to 9.38 and improves the R-precision metric from 0.325 to 0.341. 
\begin{table}[t] \small
	\centering
	
	\resizebox{1\linewidth}{!}{
		\begin{tabular}{l|c|c|c|c|c}\toprule
			Model                                      &Type              &Param [B]         &Data [M] &FID $\downarrow$    &Speed [s]      \\
			\midrule
			DALL-E               & AR               & 12                     & 250                  & 27.5     & -          \\
			
			CogView2         & AR               & 6                      & 30                   & 24.0     &  -          \\ 
			Muse-3B &AR & 3&5100&7.88&1.82\\
			Parti-350M          & AR               & 0.35                   & \textgreater800      & 14.10   &5.25          \\ 
			Parti-3B    & AR               & 3                    & \textgreater800      & 8.10          & 8.12      \\ 
			\midrule
			
			GLIDE           & DF               & 5                      & 250                  & 12.24             &    19.20\\
			
			Imagen   & DF               & 7.9                    & 860                  & 7.27                & 10.86\\ 
			DALL·E 2    & DF               & 6.5                    & 250                  & 10.39              &   -\\
			eDiff-I       & DF               & 9.1                    & 1000                 & 6.95              &   36.44\\ 
			LDM             & DF               & 1.45                   & 400                  & 12.63             &   4.83 \\
			
			\midrule
			
			LAFITE    & GAN              & 0.23            & 3                    & 26.94            &   0.03  \\
			GiGaGAN & GAN & 1.0 &980& 9.09& 0.15\\
			GALIP                          & GAN              & 0.32              & 12                   & 12.54     & 0.04 \\
			StyleGAN-T & GAN &1.2 &250 &13.90&0.10\\
			TIGER (ours) &GAN & 0.34 & 12 & 11.96 & 0.04\\
			\bottomrule
	\end{tabular}}
	\caption{We compare the performance of large pretrained autoregressive models (AR), diffusion models (DF), and GANs under zero-shot setting on the COCO test dataset.}
	\label{table2}
	\vspace{-0.4cm}
\end{table}

As shown in Tab. \ref{table2}, we conduct the quantitative comparison between our proposed TIGER and influential methods on zero-shot text-to-image synthesis task, including DALL-E \cite{ramesh2021zero,ramesh2022hierarchical}, CogView2 \cite{ding2022cogview2}, Muse \cite{chang2023muse}, GLIDE \cite{nichol2021glide}, eDiff \cite{balaji2022ediffi}, GiGaGAN \cite{kang2023scaling}, and StyleGAN-T \cite{sauer2023stylegan} \textit{etc. } Compared with autoregressive models (AR) and diffusion models (DF), our TIGER achieves competitive performance with faster speed, smaller training data size and fewer model parameters. Especially, compared with LDM (DF), TIGER obtains slight performance improvement (FID: 11.96 \textit{vs.} 12.63) with 3\% training data size, $\sim$23\% model parameters and $\sim$120$\times$ inference speed; compared with Parti-350M (AR), TIGER obtains decrease FID from 14.10 to 11.96 with less training data size and $\sim$130$\times$ inference speed. Besides, compared with other GAN methods, our TIGER achieves outstanding results. 
\subsection{Qualitative Evaluation}
As shown in Fig. \ref{nn5-1}, we conduct the qualitative comparison between our proposed TIGER, the single-stage method DF-GAN \cite{tao2022df} and the stacked method AttnGAN \cite{xu_attngan_2018}. Compare with other works, our generated images are more photo-realistic and text-matching. For example, in the 1$^{st}$ column, given the text ``A blue and red train leaving the station on a sunny day", our TIGER generates the desired images which match the attributes ``blue and red", ``train", ``station", and ``sunny". While the image generated by AttnGAN doesn't meet the requirements seriously, and the image generated by DF-GAN only saw the red and blue train and doesn't meet other attributes. In the 6$^{th}$ column, given the text ``This orange and black small bird has a straight pointed beak", the image generated by TIGER has all the mentioned attributes. However, the image generated by DF-GAN does not reﬂect ``straight pointed beak" and the image generated by AttnGAN is a little blurry. In the 8$^{th}$ column, TIGER produces the desired image, but other methods don't produce clear enough images to meet all attributes. 

\begin{figure}[t]
	\setlength{\abovecaptionskip}{0.15cm}
	
	\centerline{\includegraphics[width=\linewidth]{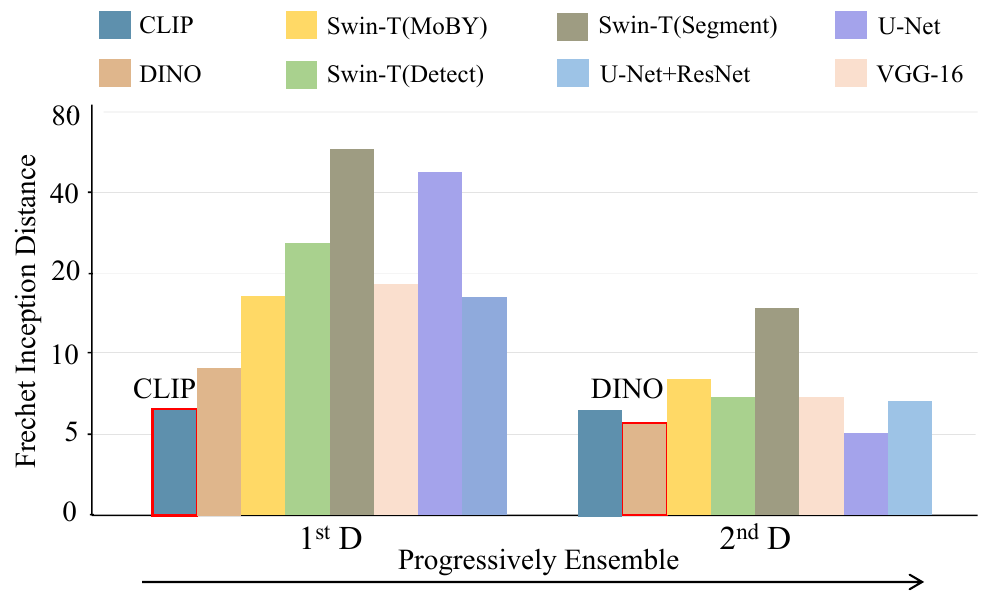}}
	\caption{Ablation experiments of different pretrained vision models. We progressively ensemble the pretrained vision models and select the best performing model (lower FID on COCO test set) in each round. At last, we select CLIP and DINO sequentially.}
	\label{nn6}
\end{figure}
\subsection{Ablation Study}
As shown in Tab. \ref{table3}, to verify the superiority of each component in our proposed TIGER, we deploy our experiments on the COCO test set \cite{lin2014microsoft}. These components include pretrained vision models, adapter, global fusion module and CLIP and DINO Layer Selection. The VE-D stands for our vision-empowered discriminator and the GFM stands for global fusion module.

\noindent{\bf Baseline.} The baseline is our proposed a single-stage text-to-image GAN \cite{tao2020df}, which has a CNN-based generator/discriminator and is trained from scratch.

\noindent{\bf Pretrained Vision Models.} As shown in Fig. \ref{nn6}, we conduct the ablation experiments of different pretrained vision models, include: VGG \cite{simonyan_very_2015}, Swin-T (Moby, Segment, Dectet) \cite{liu2021swin}, CLIP \cite{radford2021learning} , DINO \cite{caron2021emerging}, U-Net \cite{schonfeld2020u}, \textit{etc.} We experimentally found that multi-level pretrained representations (DINO, CLIP) can significantly improve model performance. In the 1$^{st}$ round, we add the first pretrained representation to the discriminator, which means our vision-empowered discriminator only has one sub discriminator. The CLIP model performs best in the 1$^{st}$ round.  In the 2$^{nd}$ round, we keep the selection from the previous round unchanged and add the second pretrained vision model to our discriminator. Next, the DINO model is selected for our model. In the 3$^{rd}$ round, our model trains slowly and fails to converge, perhaps due to our limited computing resources. So, we only add two pretrained models (CLIP, DINO).

\noindent{\bf Adapter.} To maximize our model performance, we experimented with two different adapter architectures (shown in Fig. \ref{nn45} (b) and (c)). The adapter A and B are used to process multi-level representations (DINO, CLIP). Experiments show that different adapter architectures have an important impact on model performance, and the best adapter pair decreases FID from 7.94 to 5,48 and improves R-precision from 0.308 to 0.348. At last, we equipped first sub discriminator (CLIP) with adapter A and equipped second sub discriminator (DINO) with adapter B.
\begin{table}[!t] \small
	\centering
	\resizebox{0.95\linewidth}{!}{
		\begin{tabular}{l|c|c}\toprule
			Architecture                                              &FID $\downarrow$             &CS $\uparrow$ \\ \midrule
			Baseline                                                  & 19.32                      & 0.278         \\
			+ HFBlock                                     & 11.52                        & 0.314           \\
			+ Semantic Contrastive Loss                               & 8.98                      & 0.329          \\
			+ VE D (CLIP)                                     & 5.94                        & 0.322          \\
			+ VE D (CLIP, DINO)                                & \textbf{5.48}                      & \textbf{0.348 }          \\
			\midrule
			TIGER w/o GFM & 9.74                  & 0.302 \\
			GFM w/o 3$\times$3DW-Conv & 8.91                  & 0.326 \\
			GFM w/o 1$\times$1 Conv & 6.24                  & 0.331 \\
			GFM w/o 5$\times$5 Dilated Conv& 7.59                  & 0.329 \\
			GFM w/o 3$\times$3 DW-Conv, 1$\times$1 Conv& 7,72                  & 0.318 \\
			\midrule
			TIGER w/ Adapter (A, A)                          & 5.96                  & 0.333           \\
			TIGER w/ Adapter (B, B)                          & 6.84                  & 0.329           \\
			TIGER w/ Adapter (B ,A)                          & 7.94                  & 0.308           \\
			TIGER w/ Adapter (A ,B)                  & \textbf{5.48}                  & \textbf{0.348}           \\
			\midrule
			TIGER w/ CLIP (2$^{nd}$, 5$^{th}$ )                           & 6.85         & 0.335        \\
			TIGER w/ CLIP (1$^{st}$, 5$^{th}$, 9$^{th}$)                           & 10.42         & 0.336   \\
			TIGER w/ CLIP (2$^{nd}$, 5$^{th}$, 9$^{th}$)                       & \textbf{5.48}         & \textbf{0.348}          \\
			TIGER w/ CLIP (2$^{nd}$, 5$^{th}$, 9$^{th}$, 12$^{th}$)                       & 7.96         & 0.321         \\
			\midrule
			TIGER w/ DINO (1$^{st}$, 5$^{th}$)                          & 11.40                  & 0.311         \\
			TIGER w/ DINO (2$^{nd}$, 5$^{th}$)                          & 9.47                  & 0.328         \\
			TIGER w/ DINO (1$^{st}$, 5$^{th}$, 9$^{th}$)                          &  \textbf{5.48}                  & \textbf{0.348}          \\
			TIGER w/ DINO (1$^{st}$, 5$^{th}$, 9$^{th}$, 12$^{th}$)                       & 13.97         & 0.319         \\
			\bottomrule
	\end{tabular}}
	\caption{The performance of different components of our model on the test set of COCO.}
	\label{table3}
	\vspace{-0.4cm}
\end{table}

\noindent{\bf Global Fusion Module.} We ablate the operations in our proposed global fusion module. First, we try to delete the global fusion module in our TIGER but get worse performance, which indicates the global fusion module can enhance our model capacity. Besides, we conduct experiments to verify the effectiveness of 3$\times$3 depth-wise convolutional network, 1$\times$1 convolutional network and 5$\times$5 dilated convolutional network. The ablation experiments confirm they play a positive role in our global fusion module. 

\noindent{\bf CLIP and DINO Layer Selection.} Inspired by GALIP \cite{tao2023galip}, different layers in CLIP/DINO have performance impact. Generally, first few layers contain some low-level general visual features, while last few layers contain some low-level abstract concepts \cite{kumari2022ensembling}. To further improve performance, we experimented with various layer combination strategies. In the end, we chose the 2$^{nd}$, 5$^{th}$, 9$^{th}$ layer of CLIP and the 1$^{st}$, 5$^{th}$, 9$^{th}$ layer of DINO.
\section{Conclusion}
In this paper, we propose TIGER, a text-to-image GAN with pretrained representations, which aimed to build a more powerful and faster text-to-image model.  To be specific, we propose a vision-empowered discriminator and a high-capacity generator. The vision-empowered discriminator absorbs the complex scene understanding ability and domain generalization ability from pretrained vision model to enhance the performance. And the high-capacity generator aims to achieve effective text-image fusion while increasing the model capacity. The high-capacity generator consists of multiple novel high-capacity fusion blocks (HFBlock). Extensive experiments demonstrate the outstanding performance of our proposed TIGER. In the future, we hope to try to pretrain TIGER on a larger dataset to further enhance the performance of zero-shot text-to-image synthesis.

\bibliographystyle{named}
\bibliography{ijcai24.bib}

\end{document}